\documentclass{article}

\usepackage{arxiv}

\usepackage[utf8]{inputenc} 
\usepackage[T1]{fontenc}    
\usepackage{hyperref}       
\usepackage{url}            
\usepackage{booktabs}       
\usepackage{amsfonts}       
\usepackage{nicefrac}       
\usepackage{microtype}      
\usepackage{lipsum}
\usepackage{graphicx}
\usepackage{amsmath}
\usepackage{bm}
\graphicspath{ {./images/} }
\usepackage[round]{natbib}

\title{The impossibility theorem of machine fairness \\ a causal perspective}

\author{
 Kailash Karthik Saravanakumar \\
  Columbia University \\
  New York NY 10027 \\
  \texttt{kailashkarthik.s@columbia.edu} \\
}

\pdftrailerid{} 
\begin{document}
\maketitle
\begin{abstract}
With the increasing pervasive use of machine learning in social and economic settings, there has been an interest in the notion of machine bias in the AI community. Models trained on historic data reflect biases that exist in society and propagated them to the future through their decisions. There are three prominent metrics of machine fairness used in the community, and it has been shown statistically that it is impossible to satisfy them all at the same time. This has led to an ambiguity with regards to the definition of fairness. In this report, a causal perspective to the impossibility theorem of fairness is presented along with a causal goal for machine fairness.
\end{abstract}


\section{Introduction}
The 21st century has seen the development of machine learning from a purely scientific pursuit to a ubiquitous commodity affecting every human life. Its implications have largely been positive, with many socially useful and impacting applications being developed that leverage machine learning techniques. But recent studies have revealed that data-driven classifiers may not conform to social norms of fairness and bias. Thus, the use of such models in socially and economically critical applications ranging from crime recidivism prediction, hiring decision recommendation to loan provision recommendation can have a large (and possibly adverse) impact on the society. It has been shown that, if left unattended, data-driven models can propagate any pre-existing bias towards disadvantaged sections of society, sometimes even amplifying it.

A report published by ProPublica \citep{propublica} revealed that the automated recidivism prediction system COMPAS was biased against the African-American community in its predictions. This report led to three metrics of fairness becoming ubiquitous in the literature and the AI community - Demographic Parity, Equalized Odds \citep{eqodds} and Predictive Parity \citep{fairness-metrics-explained}. \cite{impossibility} proved statistically that theze metrics are mutually incompatible and no more than one metric can be satisfied at a time (with the exception of certain trivial conditions).

The discrepancy of these metrics has led to the lack of an agreed upon definition of fairness or recommendations on when one metric might be applicable over the others. The literature has also deemed fairness as a trade-off to accuracy of the classifiers \cite{tradeoff}. 

Unfairness in a data-driven computational models originates from exposure bias - being exposed to a certain type of data distribution that is at odds with the expected behaviour of the model itself. Thus, imposing fairness as a goal of a machine learner could be seen as a trade-off to performance because a socially fair classifier would need to deviate from the underlying data distribution and introduce fairness to its predictions. However, one must note that the presence of machine bias can be either necessary or adversarial depending on the scenario.  For instance, a sentiment prediction model’s bias towards words like ”despair”, ”tragedy” to predict text sentiment is justified, and more importantly desirable.  However, a model’s bias towards a social class like a certain race or gender can be undesirable, especially when the bias is against the accepted norms of social fairness. Such attributes or variables that need to be protected from machine bias are called \textit{sensitive} or \textit{protected} attributes. \\

The main contributions of this report are:
\begin{enumerate}
    \item We present a causal perspective on the impossibility theorem and show that there can not exist a data generation process that can satisfy the three metrics of machine fairness
    \item We present an alternate representation of the goal of fairness from a causal perspective and question the interpretation of fairness as a trade-off to accuracy
\end{enumerate}

\section{Notation}

For simplicity of notation, we consider the task of binary prediction without loss of generality. Let an instance of the underlying data distribution $x_i$ be used by a machine classifier for prediction of the label $y_i$. These random variables are drawn from the distributions $x_i \in \bm{R^d}$ and $y_i \in \bm{Y} = \bm{\{0, 1\}}$. Let $a$ be the protected attribute as defined above. The variable is drawn from the distribution $a \in \bm{A}$. The goal of a predictor is to learn a function $f: \bm{X} \rightarrow{} \bm{Y}$ that approximates the true underlying joint distribution $\bm{X} \times \bm{Y}$. The model is then used to make predictions $\hat{y}$ on unseen data instances $x_i$ such that it approximates their true labels $y_i$ well. 
\begin{equation*}
    \text{Let }P_a[\hat{y}] := P [\hat{y} \mid A = a].
\end{equation*}

\section{Metrics of Machine Fairness}
\label{sec:metrics}

Though there have been multiple metrics of fairness proposed, the three commonly used ones are presented below. For a more thorough survey of all the metrics of fairness, please refer to \citep{survey}.

A model is said to calibrated or fitted if it approximates the underlying joint distribution well. In such cases, there exists a strong correlation between the true labels $y$ and the predictions $\hat{y}$. This means that for a well calibrated classifier, $Y \not\!\perp\!\!\!\perp \hat{Y}$. For all the analyses done in this report, we consider well calibrated classifiers only.

A sensitive attribute can be considered as a source of machine bias only if it has a correlation with the true label. If the true label was indifferent to the sensitive attribute, then the presence or absence makes no difference to both the true label and the prediction. Thus, for this analysis, we only consider sensitive attributes that are correlated with the true labels  $A \not\!\perp\!\!\!\perp Y$ and thus can introduce machine bias if the learning algorithm is not equipped to handle possible bias.

\subsection{Demographic Parity}

Statistical Implication - $\hat{Y} \perp\!\!\!\perp A$

This metric ensures that the predictions are independent of the sensitive attribute.  Thus prediction probabilities are equal across all values of the sensitive attribute, thus preventing the model from having disparate prediction bias towards a certain label for any sensitive group.

\subsection{Equalized Odds}

Statistical Implication - $Y \perp\!\!\!\perp A \mid \hat{Y}$

This metric ensures that the accuracy of the model is not dependent on the sensitive attribute value. Thus, the probability of predictions are independent of the the sensitive attribute for each target label groups. This encourages the model to be faithful to the underlying distribution of sensitive attribute across the target groups.

\subsection{Predictive Parity}

Statistical Implication - $\hat{Y} \perp\!\!\!\perp A \mid Y$

This metric ensures that the calibration of the model is not dependent on the sensitive attribute value. Thus, the probability of correctness of a prediction is the same for all values of the sensitive attribute. This prevents models from being biased towards making incorrect predictions for any sensitive group.

\section{Impossibility Theorem}
\label{sec:impossibility}
    
The Impossibility Theorem \cite{impossibility} states that no more than one of the three fairness metrics of demographic parity, predictive parity and equalized odds can hold at the same time for a well calibrated classifier and a sensitive attribute capable of introducing machine bias.

In this section, we present a causal explanation of the above statement from a data-generating perspective that leverages intuitions of causal inference \cite{pearl2009causality}. We assume that the classifier is well calibrated and the sensitive attribute is correlated with the true labels. This entails that the causal graph corresponding to the data generation process will always have an unblocked path between $Y - \hat{Y}$ and $Y - A$.

D-Separation \cite{pearl2009causality} is a criterion for the identification of conditional independence relationships from causal diagrams. \textit{Blocking} is the prevention of flow of probabilistic distributional influences between two nodes in the diagram. Directed paths between any two nodes in a causal diagram can be decomposed into a sequences of triplets. There are three types of triplets - causal chain, common cause and common effect. While the first two types are unblocked when none of the nodes in the triplets are observed, common effects are unblocked when the common effect node or any of its descendants are observed.

If all the paths between two variables are composed of only active triplets, then the variables are dependent on each other, conditioned on any node that is observed to make the paths active. Such variables are said to be \textit{d-connected}. If all the paths between the nodes are blocked, either by observed nodes or through common effect colliders, then the nodes are (conditionally) independent and are said to be d-separated.

In the following causal diagrams, let the curved lines denote an unblocked path of arbitrary length and configuration. The presence of the curved line implies that the nature of the intervening path is not relevant for this analysis as long as it is unblocked. We also ignore all the attributes in $X$ apart from the sensitive attribute $A$. The presence and configuration of the other attributes are not relevant to the analysis of fairness and do not change any of the results.

\subsection{Demographic Parity}
    
The causal diagram of a data generation process corresponding to a classifier that satisfies demographic parity is shown below. It can be observed that $Y$ and $\hat{Y}$ are d-connected through the unblocked represented by the curved line. Similarly, $Y$ and $A$ are d-connected. Thus the assumption that the model is well calibrated and that the sensitive attributed is correlated to the true label are satisfied. It can also be observed that the sensitive attribute is d-separated from the predicted label since the path contains the collider through the node $Y$.

This is the only data generation process configuration that satisfies the assumptions on the model as well as demographic parity.

\begin{figure}[h]
    \centering
    \includegraphics[width=0.5\linewidth]{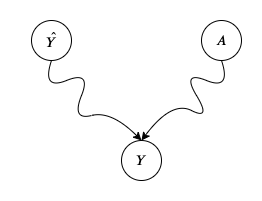}
    \caption{Causal Diagrams for Demographic Parity}
    \label{fig:dem_parity}
\end{figure}

In this generative process, $\hat{Y}$ is not a d-separating set for $A$ and $Y$. Thus, observing $\hat{Y}$ does not make $A$ and $Y$ independent and thus predictive parity can not hold in such a process.

It can also be seen that observing Y opens up the collider on the node which makes the sensitive attribute and the prediction d-connected. Thus, equalized odds can not hold true in the model corresponding to this diagram.

\subsection{Equalized Odds}
    
The causal diagrams for a model of data generation that satisfies the equalized odds notion of fairness as well as the model assumptions are shown below. There are three possible configurations depending the direction of the edges entering/leaving $Y$. If the node $Y$ is the center of either a causal chain or a common cause triplet, then equalized odds holds as the observation of $Y$ blocks the path between $\hat{Y}$ and $A$.

\begin{figure}[h]
    \centering
    \includegraphics[width=0.5\linewidth]{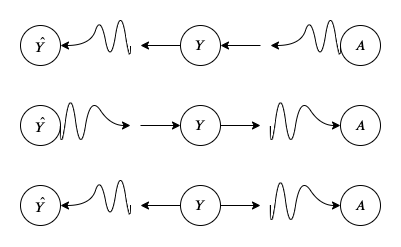}
    \caption{Causal Diagrams for Equalized Odds}
    \label{fig:eq_odds}
\end{figure}

It can be observed from the causal diagram that the path between $A$ and $\hat{Y}$ is unblocked unless $Y$ is observed. This entails that demographic parity can not hold in such a model.

The prediction label $\hat{Y}$ does not lie between $Y$ and $A$ and thus, its observation hence has no effect on their d-connectedness. Thus, predictive parity can not hold in such settings.
    
\subsection{Predictive Parity}
    
The data generation process for a model that satisfies both the model assumptions as well as predictive parity is shown below. As mentioned in the equalized odds scenario, there are three possible configurations depending on the directions of the causal relationships of $\hat{Y}$ on the path between $A$ and $Y$.

\begin{figure}[h]
    \centering
    \includegraphics[width=0.5\linewidth]{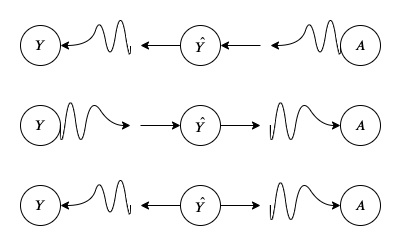}
    \caption{Causal Diagrams for Predictive Parity}
    \label{fig:pred_parity}
\end{figure}

The causal diagram is very similar to the equalized odds scenario. In the same way, the sensitive attribute is always d-connected to the prediction and thus demographic parity can not be satisfied by such a model.

In addition to this, it can be observed that $Y$ can never d-separate $A$ and $\hat{Y}$ as it does not lie on the unblocked causal paths between them and hence can not block them. It can thus be seen that equalized odds can not hold in such a model.

\subsection{Inferences}

From these causal diagrams, it can be concluded that a computational model that satisfies one of the three notions of fairness can not simultaneously hold any of the other two. This inability can not be attributed to the lack of capability of the models but rather  should be attributed to the restrictions of the data generation regime. We have proved, by being classifier model-agnostic, using causal techniques that the impossibility theorem is true.

Thus, a machine learning practitioner building computational models for socially impacting applications must choose one of these metrics to satisfy at the expense of the other two. However, the issue with such a selection is that any model that is deemed to be unfair by any of these metrics can be disputed by citing one of the other two metrics that can be satisfied.

Thus, a refinement of metrics for fairness is required, taking into consideration the underlying causal structure. One possibility is look at fairness at an individual level using counterfactuals \cite{counterfactual}. However, there has been criticism that the notion of counterfactual fairness requires a lot of constraints for identifiability. In order to satisfy counterfactual fairness, it is stated that all the descendants of the sensitive attribute should be avoided as features for classification. But in reality, most sensitive attributes are parent-less nodes whose descendants span the entire causal graph which makes it hard to satisfy this constraint.

An alternate methodology would be one that uses the notion of causal graphs and is proposed in the next section.

\section{Fairness using Causal Diagrams}

The reason there is a lack of fairness in many data-driven classifiers is due to the discrepancy between what is deemed correct by the data and what is thought of as "correct" by the society. Many papers in the literature have mentioned that fairness in machine learning is not a statistical but rather a sociological issue in using machine learnt classifiers in practical settings. The exposure bias of machine learners is at odds with what is expected of a \textit{desirable} classifier which makes these models unfair.

The objective of a learning algorithm is usually empirical risk minimization (ERM), which tries to bridge the gap between the true labels and the predicted labels. Models are penalized for deviating from the underlying data distribution. We argue that the notion of fairness is orthogonal. A fair classifier is one that bridges the gap between the fair labels and the predicted labels. Such hypothetical classifiers try to negate the historical dependence of the true label on the sensitive attribute by making predictions that take into consideration the sensitivity of certain attributes - thereby deviating from the true labels in the (historic) data. 

The goal of the learning algorithm can then be thought of as manipulating the past data and making it consistent with a notion of fairness in order to make fair predictions in the future. This is a way to ensure fairness as classification models usually assume that data is independent and identically drawn from the same distribution. Conceptually, this can be thought of as introducing a correction to the labels in the dataset.
    \begin{align*}
        f_{\text{ERM}} &= \text{argmin}_{f} [\bm{P}_{(x_i, y_i) \sim \bm{D}}(y_i \neq \hat{y_i})] \\
        f_{\text{fair}} &= \text{argmin}_{f} [\bm{P}_{(x_i, y_i) \sim \bm{D}}(y_i^{\text{fair}} \neq \hat{y_i})] \\
        \text{where } y_i^{\text{fair}} &= f_c(y_i)
    \end{align*}

The true labels from the data are determined to be unfair based on some socially acceptable correction criterion and changed using the corresponding correction function $f_c$.

The correction function introduce a notion of randomization that depends on the value of the sensitive attribute, trying to disentangle the dependence between the true labels found in data and the desired labels from the classifier. Thus, the goal of the learning algorithm is no longer to produce a model that maximizes the fit of the training data, but rather to fit the training data with this correction function in the picture.

    \begin{equation*}
        Y_{\text{goal}} \neq Y_{\text{ERM}}
    \end{equation*}

Thus, a desirable classifier is no longer constrained the requirement for perfect calibration ($Y \perp\!\!\!\perp?\ \hat{Y}$). The only constrain then is the fact that the sensitive attribute has some effect of the true labels in the dataset ($Y \not\!\perp\!\!\!\perp A)$.

\subsection{Conceptual Realization of Correction}

Under this constraint, the causal diagram for the data generation process can then be represented by introducing a correction variable $C$ that blocks the influence of $Y$ on $\hat{Y}$. That is, the correction variable will determine if information from the true label is to be propagated to the prediction or not. The correction variable by itself is a function of the protected attribute ($C \not\!\perp\!\!\!\perp A)$ but does not let information from $A$ pass through it. While the flow of information from $Y$ to $\hat{Y}$ doesn't have to be necessarily blocked for the advantaged classes, this is desired for the disadvantaged class to ensure fairness.

\begin{figure}[h]
        \centering
        \includegraphics[width=0.6\linewidth]{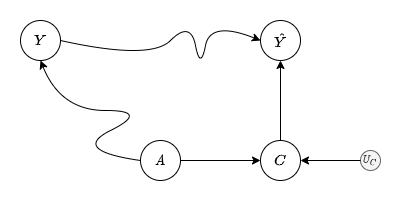}
        \caption{Causal Diagrams with Correction Variable}
        \label{fig:correction}
    \end{figure}

    \begin{figure}[h]
        \centering
        \includegraphics[width=0.45\linewidth]{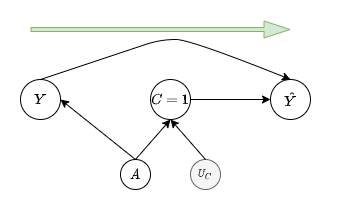}
        \includegraphics[width=0.45\linewidth]{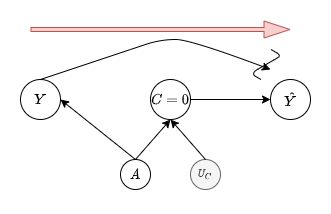}
        \caption{Effect of Correction Variable}
        \label{fig:effect_correction}
    \end{figure}
    
    The structural equation for $\hat{Y}$ can be thought to be of the form
    \begin{align*}
        f_c &= A \oplus U_C \\
        \hat{Y} &= \begin{cases}
      g_{\text{classifier}}(y), & \text{if}\ c=1 \\
      g_{\text{fairness}}(a), & \text{if}\ c=0
    \end{cases}
    \end{align*}

Thus, a machine learning practitioner can introduce a debiasing correction of her/his choice by choosing the degree of deviation from the data-driven dependence of $Y$ on the features. One could think of this correction as being similar to the off-policy $\epsilon$-greedy strategy in reinforcement learning. The probability with which $U_C$ takes 0 or 1 can then be a hyperparameter tuned to reflect (1) how unfair the data is; and (2) how much of a deviation from historic data is desired in the classifier developed.

\subsection{Existing Fairness Metrics using Correction Variables}

\subsubsection{Demographic Parity}
    
In the new causal mechanism, though $A$ is dependent on $\hat{Y}$, the influence is only through the activation or disabling of the gate $C$. Thus, we achieve a looser version of demographic parity where the sensitive attribute does not have any direct causal path to the prediction and all the paths are through the correction variable $C$.
    
\subsubsection{Equalized Odds}
    
In this scenario as well, conditioned on the true label $Y$, all the causal paths from the sensitive attribute to the prediction are through the correction variable and there exists no other influence.
    
\subsubsection{Predictive Parity}
    
In the new scenario, predictive parity can never be achieved as there is always a causal influence of the sensitive attribute on the true label $Y$. I argue that this metric should not be satisfied as it is against the intuition we developed earlier about the desired classifier. The introduction of the correction mechanism was to selectively offset the true labels in the dataset with the aim of negating existing bias. Thus, a fair classifier is one that would preferentially reverse the true label $Y$ for the disadvantaged class $A = a$ that has been subject to prejudice. Thus, we would expect that the $P(Y=v \mid \hat{Y}=v, A=a) > P(Y=v \mid \hat{Y}=v, A=d)$, where $v$ is any value in the range of $Y$ and $\hat{Y}$, $a$ is the advantaged group and $d$ is the disadvantaged group.

\subsubsection{Modified Fairness Equations}

In the new regime which includes the correction variable, the notions of demographic parity and equalized odds can be satisfied together conditioned on the correction.
\begin{align*}
    \hat{Y} &\perp\!\!\!\perp A \mid (C=0) \\
    \hat{Y} &\perp\!\!\!\perp A \mid Y, C
\end{align*}

Here we achieve a trade-off between satisfying demographic parity and equalized odds by tuning the level of a hyperparameter that determines the probability that $C$ takes values 0 or 1.

\section{Conclusion}

In this report, a proof for the impossibility theorem of fairness was presented from a causal inference perspective. It was shown that the three contemporary metrics defy the possibility of a single data generation process satisfying them all and thus will always be at odds with each other. 

An explanation on why the ERM solution obtained by training classifiers on data might not be desirable from a fairness perspective was provided followed by a theoretical proposal of a correction to the training data. This correction seeks to offset the bias present in the dataset and align it with current notions of fairness (which themselves might evolve with time).

\bibliographystyle{plainnat}
\bibliography{references}  


\end{document}